\documentclass[a4paper,11pt]{article}

\usepackage{amsmath}
\usepackage{array} 
\usepackage{amssymb} 
\usepackage{bm} 
\usepackage{bclogo}
\usepackage{booktabs} 
\usepackage{calc}	  
\usepackage{caption} 
\usepackage{enumitem}
\usepackage{diagbox} 
\usepackage{float}
\usepackage{graphicx}  
\usepackage[margin=25mm]{geometry}
\usepackage[colorlinks=true,allcolors=blue!60!black,breaklinks]{hyperref} 
\usepackage{listings} 
\usepackage{mymacros} 
\usepackage{multicol} 
\usepackage{multirow} 
\usepackage{nicefrac}
\usepackage{natbib} 
\usepackage{pifont} 
\usepackage{ragged2e}    
\usepackage{subcaption}
\usepackage{syntax}
\usepackage{transparent}
\usepackage{textcomp} 
\usepackage{times} 
\usepackage{tcolorbox} 
\usepackage{colortbl}   
\usepackage[textwidth=2.1cm]{todonotes} 
\usepackage[normalem]{ulem} 
\usepackage{url}       
\usepackage{pifont}

\newcommand{\drgvar}[1]{\textbf{#1}}
\newcommand{\ttboxlab}[1]{\texttt{#1}}
\newcommand{\strout}[1]{\sout{\mbox{#1}}}
\newcommand{\wsp}{\hspace{4.5mm}}
\newcommand{\posn}[2]{#1\kern-0.15em.\kern-0.15em#2}
\newcommand{\matched}[1]{\textcolor{green!40!black}{#1}}
\newcommand{\nonmatched}[1]{\textcolor{red!70!black}{#1}}
\newcommand{\ntt}[2][1]{\textsmaller[#1]{\texttt{#2}}}
\newcommand{\pair}[1]{\ensuremath{\langle#1\rangle}}
\definecolor{pmbBlue}{RGB}{14,163,172}
\newcommand{\clause}[2][0]{\mbox{$\langle$\ntt[#1]{#2}$\rangle$}}
\newcommand{\cross}{\textcolor{red}{\ding{54}}}
\newcommand{\cmark}{\textcolor{green}{\ding{52}}}
\newcommand{\bnfstr}[1]{\textcolor{red}{\textbf{#1}}}
\newcommand{\bnfmath}[1]{\textcolor{red}{\ensuremath{\bm{#1}}}}

\floatstyle{plaintop}
\newfloat{mybnf}{tbp}{lop}
\floatname{mybnf}{Definition}

\newcommand{\makeunderscoreletter}{\catcode`\_=11}
\newcommand{\makeunderscoreactive}{\catcode`\_=\active}
\newcommand{\mytilde}{\raise.17ex\hbox{$\scriptstyle\mathtt{\sim}$}}

\newcommand{\taclsys}{\textsc{Noord et al.18}}
\newcommand{\iwcssys}{\textsc{Noord et al.19}}
\newcommand{\liusys}{\textsc{Liu et al.}}
\newcommand{\fancellusys}{\textsc{Fancellu et al.}}
\newcommand{\evangsys}{\textsc{Evang}}
\newcommand{\tacl}{\textsc{Noord18}}
\newcommand{\iwcs}{\textsc{Noord19}}
\newcommand{\liu}{\textsc{Liu}}
\newcommand{\fancellu}{\textsc{Fancellu}}
\newcommand{\evang}{\textsc{Evang}}

\title{The First Shared Task on\\Discourse Representation Structure Parsing}
\date{}
\author{Lasha Abzianidze \qquad Rik van Noord \qquad Hessel Haagsma \qquad Johan Bos\\  
        CLCG, University of Groningen\\
       \texttt{\{l.abzianidze,\,r\!.i\!.k.van.noord,\,hessel.haagsma,\,johan.bos\}@rug.nl}
}

\begin{document}
\makeunderscoreletter
\maketitle
\thispagestyle{empty}
\pagestyle{empty}

\begin{abstract}
The paper presents the IWCS 2019 shared task on semantic parsing where the goal is to produce Discourse Representation Structures (DRSs) for English sentences.
DRSs originate from Discourse Representation Theory and represent scoped meaning representations that capture the semantics of negation, modals, quantification, and presupposition triggers.
Additionally, concepts and event-participants in DRSs are described with WordNet synsets and the thematic roles from VerbNet. 
To measure similarity between two DRSs, they are represented in a clausal form, i.e. as a set of tuples.
Participant systems were expected to produce DRSs in this clausal form.
Taking into account the rich lexical information, explicit scope marking, a high number of shared variables among clauses, and highly-constrained format of valid DRSs, all these makes the DRS parsing a challenging NLP task.
The results of the shared task displayed improvements over the existing state-of-the-art parser.  
\end{abstract}

\section{Introduction}
\label{sec:intro}

Semantic parsing has been gaining in popularity in the last few years.
There have been a series of shared tasks in semantic parsing organized, where each task requires to generate meaning representations of specific types: 
Broad-Coverage Broad-coverage Semantic Dependencies \citep{oepen-etal-2014-semeval, oepen-etal-2015-semeval},
Abstract Meaning Representation \citep{semeval-AMR1, semeval-AMR2},
or Universal Conceptual Cognitive Annotation \citep{hershcovich2019semeval}.

The Discourse Representation Structure (DRS) parsing task extends this development by aiming at producing meaning representations that (i) come with more expressive power than existing ones and (ii) are easily translatable into formal logic, thereby opening the door to applications that require automated forms of inference \citep{BlackburnBos:2005,rteBook:2013}.
DRSs are meaning representations employed by Discourse Representation Theory (DRT, \citealp{kampreyle:drt}).
They have been successfully applied for wide-coverage semantic representations \citep{Bosetal:04,step2008:boxer}, Natural Language Inference \citep{BosMarkert2005EMNLP,Bjerva2014SemEval}, and Natural Language Generation \citep{BasileBos2013ENLG}. 
To the best of our knowledge, there has never been a shared task on scoped meaning representations.

The aim of the task is to compare semantic parsing methods and the performance of systems that take as input an English text and provide as output the scoped meaning representation of that text 
Since a DRS combines logical (negation, quantification and modals), pragmatic (presuppositions) and lexical (word senses and thematic roles) components of semantics in a single meaning representation,
the DRS parsing task shares parts of the following NLP tasks: semantic role labeling, reference resolution, scope detection, named entity tagging, word sense disambiguation, predicate-argument structure prediction, and presupposition projection.

There are only a few previous approaches to DRS parsing. Traditionally, due to the complexity of the task, it has been the domain of symbolic and statistical approaches \citep{step2008:boxer, le:12, boxer}. Recently, however, neural sequence-to-sequence systems achieved impressive performance on the task \citep{neural_drs_gmb:18, drstacl:18}, without relying on any external linguistic resources.

In the first shared task on DRS parsing, taking into account the information-rich and complex structure of the target meaning representation, we tested participant systems mainly on short, open-domain English texts.
In this way, we lowered the threshold for participation to encourage higher results in the shared task and mitigate challenges associated to semantic parsing long texts.
In total five systems participated in the shared task.
The top-ranked systems outperformed the existing state-of-the-art system in DRS parsing. 
The shared task was hosted on CodaLab.%
\footnote{\url{https://competitions.codalab.org/competitions/20220}}.

\begin{figure}
\centering
\begin{minipage}{.42\textwidth}
\centering
\parbox{0pt}{
\begin{tabbing}
12\=345\=\kill
{\sc system input}:\\[0pt] 
\>\>Tom isn't afraid of anything.
\\[1mm]
{\sc system output}:\\[0pt] 
\>\>\textsmaller[1]{\texttt{%
 \begin{tabular}[t]{@{\,}l}
  \drgvar{b1} REF \drgvar{x1}\\
  \drgvar{b1} male "\posn{n}{02}" \drgvar{x1}\\
  \drgvar{b1} Name \drgvar{x1} "tom"\\
  \drgvar{b2} REF \drgvar{t1}\\
  \drgvar{b2} EQU \drgvar{t1} "now"\\
  \drgvar{b2} time "\posn{n}{08}" \drgvar{t1}\\
  \drgvar{b2} NOT \drgvar{b3}\\
  \drgvar{b3} REF \drgvar{s1}\\
  \drgvar{b3} Time \drgvar{s1} \drgvar{t1}\\
  \drgvar{b3} Experiencer \drgvar{s1} \drgvar{x1}\\
  \drgvar{b3} afraid "\posn{a}{01}" \drgvar{s1}\\
  \drgvar{b3} Stimulus \drgvar{s1} \drgvar{x2}\\
  \drgvar{b3} REF \drgvar{x2}\\
  \drgvar{b3} entity "\posn{n}{01}" \drgvar{x2}\\
 \end{tabular}}}
\\[1mm]
{\sc box format}:\\[3pt] 
\begin{drstree}{-2}{4mm}{4mm}
[{\pdrs{b0}{$t_1$}{
    \textlarger[3]{$\neg$}
      \pdrs[c]{b3}{$s_1$ ~ $x_2$}{
        afraid$\sym{.a.01}(s_1)$\\
        \wsp$\sym{Time}(s_1, t_1)$\\ 
        \wsp$\sym{Stimulus}(s_1, x_2)$\\
        \wsp$\sym{Experiencer}(s_1, x_1)$\\
        entity$\sym{.n.01}(x_2)$}\\
    $\sym{time.n.08}(t_1)$\\   
    \wsp$t_1 = \sym{now}$
    }}
    [{\pdrs{b1}{$x_1$}{
        $\sym{male.n.02}(x_1)$\\
        \wsp$\sym{Name}(x_1, \text{tom})$}}
    ]
]
\end{drstree}
\end{tabbing}
}\vspace{-7mm}
\caption{The DRS parsing task: the~system input is a short text (the PMB document \href{http://pmb.let.rug.nl/explorer/explore.php?part=99&doc_id=2308&type=der.xml}{99/2308}), and the expected output is a DRS in clausal form. Its standard visualisation in box-notation, following DRT, is presented below.}
\label{fig:drs:afraid}
\end{minipage}%
\qquad
\begin{minipage}{.53\textwidth}
\hspace{-4mm}
\scalebox{1.02}{\begin{tabular}[t]{c}
\mbox{\href{http://pmb.let.rug.nl/explorer/explore.php?part=00&doc_id=3008&type=der.xml}{00/3008}}: 
He played the piano and she sang.
\\[-1mm]
\textsmaller[1]{\texttt{
\renewcommand*{\arraystretch}{1.1}
\begin{tabular}[t]{@{}l@{\kern5mm}l@{}}\toprule
\drgvar{b6} DRS \drgvar{b1} &
\drgvar{b6} DRS \drgvar{b4}\\
\drgvar{b2} REF \drgvar{x1} &
\drgvar{b5} REF \drgvar{x3}\\
\drgvar{b2} male "\posn{n}{02}" \drgvar{x1} &
\drgvar{b5} female "\posn{n}{02}" \drgvar{x3}\\
\drgvar{b1} REF \drgvar{e1} & 
\drgvar{b4} REF \drgvar{e2}\\
\drgvar{b1} play "\posn{v}{03}" \drgvar{e1} &
\drgvar{b4} sing "\posn{v}{01}" \drgvar{e2}\\
\drgvar{b1} Agent \drgvar{e1} \drgvar{x1} &
\drgvar{b4} Agent \drgvar{e2} \drgvar{x3}\\
\drgvar{b1} Theme \drgvar{e1} \drgvar{x2} &
\drgvar{b4} Time \drgvar{e2} \drgvar{t2}\\
\drgvar{b3} REF \drgvar{x2} &
\drgvar{b4} REF \drgvar{t2}\\
\drgvar{b3} piano "\posn{n}{01}" \drgvar{x2} & 
\drgvar{b4} TPR \drgvar{t2} "now"\\
\drgvar{b1} REF \drgvar{t1} & 
\drgvar{b4} time "\posn{n}{08}" \drgvar{t2}\\
\drgvar{b1} time "\posn{n}{08}" \drgvar{t1} &
\drgvar{b6} CONTINUATION \drgvar{b1} \drgvar{b4}\\
\drgvar{b1} TPR \drgvar{t1} "now" &
\drgvar{b1} Time \drgvar{e1} \drgvar{t1}\\
\bottomrule
\end{tabular}
}}
\\\noalv{4mm}
\scalebox{1.02}{
\begin{drstree}{-1}{4mm}{5mm}
[{\psdrs{b6}{}{
    \pdrs{b1}{$e_1$ ~ $t_1$}{
        play$\sym{.v.03}(e_1)$\\
        \wsp$\sym{Time}(e_1, t_1)$\\
        \wsp$\sym{Theme}(e_1, x_2)$\\
        \wsp$\sym{Agent}(e_1, x_1)$\\
        $\sym{time.n.08}(t_1)$\\
        \wsp$t_1 \prec \sym{now}$}
    \pdrs{b4}{$e_2$ ~ $t_2$}{
        sing$\sym{.v.01}(e_2)$\\
        \wsp$\sym{Time}(e_2, t_2)$\\
        \wsp$\sym{Agent}(e_2, x_3)$\\
        $\sym{time.n.08}(t_2)$\\
        \wsp$t_2 \prec \sym{now}$}}
    {$\sym{CONTINUATION}(b_1,b_4)$}}
    [{\pdrs{b2}{$x_1$}{
        male$\sym{.n.02}(x_1)$}}
    ]
    [{\pdrs{b3}{$x_2$}{
        piano$\sym{.n.01}(x_2)$}}
    ]
    [{\pdrs{b5}{$x_3$}{
        female$\sym{.n.02}(x_3)$}}
    ]    
]
\end{drstree}
}
\end{tabular}
}
\caption{The segmented box \ttboxlab{b6} consists of a set of labelled boxes, i.e. the discourse segments \ttboxlab{b1} and \ttboxlab{b2}, and a single discourse condition.
In the condition, discourse relation holds between two discourse segments and is formatted in uppercase.
The definite noun phrase and the pronouns are presupposition (\ttboxlab{b2}, \ttboxlab{b3}, and \ttboxlab{b5}) triggers.
}
\label{fig:drs:play_piano}
\end{minipage}
\vspace{-0.4cm}
\end{figure}

\section{Task Description}
\label{sec:task}

The DRS parsing in a nutshell is presented in \autoref{fig:drs:afraid}.
Here, the input, a short English sentence, needs to be mapped to the output, a scoped meaning representation in clausal form. 
Concepts, states and events are represented by the word senses (male\sym{.n.02}, entity\sym{.n.01}, afraid\sym{.a.01}) from  WordNet 3.0 \citep{wordnet} and relations are modeled with thematic roles (\sym{Name}, \sym{Experiencer}, \sym{Stimulus}) drawn from an extended version of VerbNet \citep{Bonial:11}.

Each entity needs to introduce a discourse referent, i.e. a variable, in the right scope, form an instance of the right concepts, and be connected to other entities via thematic roles or comparison operators.
For example, in \autoref{fig:drs:afraid}, \emph{anything} introduces a discourse referent $x_2$ in the scope \ttboxlab{b3} with the help of the clause \clause{b3 REF x2}.
The clause \clause{b3 entity "\posn{n}{01}" x2} makes $x_2$ an instance of entity\sym{.n.01}.
Finally, the clause \clause{b3 Stimulus s1 x2} connects $x_2$ to the event entity $s_1$ of \emph{afraid} via the \sym{Stimulus} thematic role.   

The scopes of negation, implication, modal operators or propositional arguments need to be correctly identified. 
Proper names, pronouns, definite descriptions and possessives
are treated as presuppositions and get their own box if they cannot be resolved by the local context. Tense is locally accommodated.
For example, \autoref{fig:drs:afraid} shows how the negation operator introduces the scope (\ttboxlab{b3}) and how the named entity \emph{Tom} gives rise to the presupposition (\ttboxlab{b1}).
\autoref{fig:drs:play_piano} demonstrates how discourse segments get their own scope (\ttboxlab{b1} and \ttboxlab{b4}) and how definite noun phrases and pronouns trigger presuppositions (\ttboxlab{b2}, \ttboxlab{b3}, and \ttboxlab{b5}).
Finally, \autoref{fig:drs:his_money} depicts an implication with two scopes (\ttboxlab{b3} and \ttboxlab{b5}), modeling semantics of a universal quantifier, and nested presuppositions (\ttboxlab{b1} and \ttboxlab{b4}) due to a possessive pronoun.  

Given the aforementioned nuances of the fine-grained scoped meaning representations, the DRS parsing task represents a challenge for machine learning methods.

\section{Discourse Representation Structure} \label{sec:drs}\label{ssec:boxes}

The meaning representations used in this shared task are  based on
the DRSs put forward in DRT \citep{kampreyle:drt} and derived from the Parallel Meaning Bank \citep{PMBshort:2017}. There are some important extensions to the theory, though.
First, the DRSs are language-neutral, and all non-logical symbols are disambiguated to WordNet synsets or VerbNet roles. Furthermore, presuppositions are explicitly represented following
\citep{van_der_sandt:92} and Porjective DRT \citep{venhuizen2018discourse}. Discourse structure is analysed following by Segmented DRT \citep{asherlascarides}. As in the original DRT, DRSs are displayed in box format for reading convenience (presuppositional DRSs are displayed with outgoing arrows of the boxes that triggered them). DRSs are recursive structures, and for the purpose of evaluation, they are translated into clauses, flattening down the recursion by reification.

\begin{mybnf}[b!]
\centering
\caption{A BNF of DRSs: (possibly empty) sets are denoted with curly brackets as \{$\langle$ element$\rangle$\}.
The string elements for operators and punctuation are in red.}
\label{def:bnf}
\begin{minipage}{.8\textwidth}
\makeunderscoreactive
\setlength{\grammarparsep}{.5mm plus 1pt minus 1pt} 
\setlength{\grammarindent}{40mm} 
\footnotesize
\begin{grammar}
<DRS> ::= \{<DRS>\} <labelled BOX> 

<labelled BOX> ::= <label> <BOX>

<BOX> ::= <simple BOX> | <segmented BOX>

<simple BOX> ::= \{<discourse referent>\} \{<condition>\}

<condition> ::= <basic condition> | <complex condition>

<term> ::= <discourse referent> | <constant> 

<basic condition> ::= <semantic role> \bnfstr{(}<term>\bnfstr{,} <term>\bnfstr{)} \alt <term> <comparison operator> <term>
\alt <concept>\bnfstr{.}<pos\_sense\_number>\bnfstr{(}<term>\bnfstr{)}

<complex condition> ::= \bnfmath{\lnot}<labelled BOX> | \bnfmath{\Diamond}<labelled BOX> | \bnfmath{\Box}<labelled BOX> \alt <labelled BOX>\bnfmath{\Rightarrow}<labelled BOX>  
\alt <discourse referent>\bnfstr{:}<labelled BOX>

<segmented BOX> ::= \{<labelled BOX>\} \{<discourse condition>\}

<discourse condition> ::= <discourse relation> \bnfstr{(}<label>\bnfstr{,} <label>\bnfstr{)}

\end{grammar}
\makeunderscoreletter
\end{minipage}
\end{mybnf}

\begin{figure}[t!]
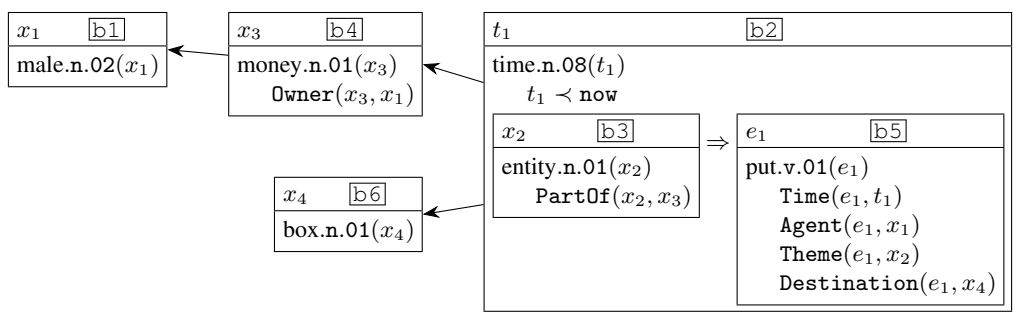

\centering
\begin{tabular}{c}
\href{https://pmb.let.rug.nl/explorer/explore.php?part=01&doc_id=2312&type=der.xml}{01/2312}: He put all his money in the box.
\\[-2mm]
\textsmaller[2]{\texttt{%
\begin{tabular}[t]{@{\,}l l @{\hspace{10mm}} l l@{\,}}
\toprule
\drgvar{b1} REF \drgvar{x1} &
    \comm{\% He [0...2] his [11...14]}
& \drgvar{b2} IMP \drgvar{b3} \drgvar{b5} &
    \comm{\% all [7...10]}
\\
\drgvar{b1} male "\posn{n}{02}" \drgvar{x1} &
    \comm{\% He [0...2] his [11...14]}
& \drgvar{b3} REF \drgvar{x2} &
    \comm{\% all [7...10]}
\\
\drgvar{b2} REF \drgvar{t1} &
    \comm{\% put [3...6]}
& \drgvar{b3} PartOf \drgvar{x2} \drgvar{x3} &
    \comm{\% all [7...10]}
\\
\drgvar{b2} TPR \drgvar{t1} "now"  &
    \comm{\% put [3...6]}
& \drgvar{b3} entity "\posn{n}{01}" \drgvar{x2} &
    \comm{\% all [7...10]}
\\
\drgvar{b2} time "\posn{n}{08}" \drgvar{t1} &
    \comm{\% put [3...6]}
& \drgvar{b4} REF \drgvar{x3} & 
    \comm{\% his [11...14]}
\\
\drgvar{b5} REF \drgvar{e1} &
    \comm{\% put [3...6]}
& \drgvar{b4} Owner \drgvar{x3} \drgvar{x1} & 
    \comm{\% his [11...14]}
\\
\drgvar{b5} Agent \drgvar{e1} \drgvar{x1} &
    \comm{\% put [3...6]}
& \drgvar{b4} money "\posn{n}{01}" \drgvar{x3} &
    \comm{\% money [15...20]}    
\\
\drgvar{b5} Theme \drgvar{e1} \drgvar{x2} &
    \comm{\% put [3...6]}
& \drgvar{b5} Destination \drgvar{e1} \drgvar{x4} &
    \comm{\% in [21...23]}
\\
\drgvar{b5} Time \drgvar{e1} \drgvar{t1} &
    \comm{\% put [3...6]}
& \drgvar{b6} REF \drgvar{x4} &
    \comm{\% the [24...27]}
\\
\drgvar{b5} put "\posn{v}{01}" \drgvar{e1} &
    \comm{\% put [3...6]}
& \drgvar{b6} box "\posn{n}{01}" \drgvar{x4} &
    \comm{\% box [28...31]}
\\
\bottomrule
\end{tabular}
}}
\\\noalv{3mm}
\begin{drstree}{-1}{8mm}{8mm}
[{\pdrs{b2}{$t_1$}{
        time$\sym{.n.08}(t_1)$\\
        \wsp$t_1 \prec \sym{now}$\\\noalv{-6pt}
         \pdrs{b3}{$x_2$}{
            entity$\sym{.n.01}(x_2)$\\
            \wsp$\sym{PartOf}(x_2, x_3)$} 
        \raisebox{-13pt}{$\Rightarrow$}
        \pdrs{b5}{$e_1$}{
            put$\sym{.v.01}(e_1)$\\
            \wsp$\sym{Time}(e_1, t_1)$\\
            \wsp$\sym{Agent}(e_1, x_1)$\\
            \wsp$\sym{Theme}(e_1, x_2)$\\
            \wsp$\sym{Destination}(e_1, x_4)$}}}
    [{\pdrs{b4}{$x_3$}{
        money$\sym{.n.01}(x_3)$\\
        \wsp$\sym{Owner}(x_3, x_1)$}}
        [{\pdrs{b1}{$x_1$}{
            male$\sym{.n.02}(x_1)$}}
        ]
    ]
    [{\pdrs{b6}{$x_4$}{
        box$\sym{.n.01}(x_4)$}}
    ]
]
\end{drstree}
\end{tabular}
\caption{The DRS contains the example of nested presuppositions triggered by the possessive pronoun \emph{his}. 
The main box \ttboxlab{b2} of the DRS presupposes a set of two DRSs.
At the same time, one of the presupposed DRSs, namely $\pair{\{\ttboxlab{b1}\},\ttboxlab{b4}}$, itself carries the presupposition \ttboxlab{b1}.
Note that the presuppositions about a male discourse referent, triggered by \emph{he} and \emph{his} separately, are merged into a single presupposition box \ttboxlab{b1}. The clauses are accompanied with aligned tokens. 
}
\label{fig:drs:his_money}
\end{figure}

A DRS always contains a main labelled box along with an optional set of presupposition DRSs (see Definition\,\ref{def:bnf}).
For example, the main labelled box in \autoref{fig:drs:afraid} is \ttboxlab{b0} while \ttboxlab{b0} is a presupposition. 
A box can be simple (e.g., the box labelled with \ttboxlab{b0} in \autoref{fig:drs:afraid}) or segmented (e.g., the box labelled with \ttboxlab{b6} in \autoref{fig:drs:play_piano}).
A simple box consists of a set of discourse referents and a set of conditions. Conditions can be basic or complex.
Basic conditions are concept predicates or relations over discourse referents and constants.
Indexicals are treated as constants, not as discourse referents \cite{indexicals:17}, for example, {\em now} is one of such indexicals (see \autoref{fig:drs:afraid}).
Complex conditions are those involving labelled boxes.
The examples of complex conditions are $\neg \ttboxlab{b3}$ in \autoref{fig:drs:afraid} and $\ttboxlab{b3}\Rightarrow\ttboxlab{b5}$ in \autoref{fig:drs:his_money}.
Finally, a segmented box contains a set of labelled boxes (\ttboxlab{b1} and \ttboxlab{4} in \autoref{fig:drs:play_piano}) and discourse conditions.
A discourse condition is a discourse relations over box labels, e.g., $\sym{CONTINUATION}(b_1,b_4)$ in \autoref{fig:drs:play_piano}.

%
\label{ssec:clauses}

The clausal form and the box-notation are two different forms of displaying scoped meaning representations \cite*{lrec:2018}.  
We consider the clausal form a machine-readable format that is suitable for the evaluation with a continuous score between 0 and 1  (see Section\,\ref{sec:evaluation}).
On the other hand, the box-notation is a human-readable format and originates from Discourse Representation Theory.
Conversion from the box-notation to the clausal form and vice versa is transparent: each box gets a label, and discourse referents and conditions in the clausal form are preceded by the label of the box they occur in.

\section{Data}
\label{sec:data}
\subsection{Released Data}
\label{ssec:rel_data}

For the shared task we released the training, development, and test data, taken from the Parallel Meaning Bank (PMB,  \citealt{PMBshort:2017}). The PMB is a parallel corpus annotated with formal meaning representations.%
\footnote{A part of the corpus can be viewed online via the PMB explorer: \url{http://pmb.let.rug.nl/explorer}}
These representations capture the most probable interpretation of a sentence; no ambiguities or under-specification techniques are employed.
The formal meaning representations are automatically constructed and manually corrected. Completely correct representations are flagged as \emph{gold}.
Representations that are partly manually corrected are marked as \emph{silver}, while the rest is marked \emph{bronze}.

The PMB release number used for the shared task is 2.2.0%
\footnote{\url{https://pmb.let.rug.nl/data.php}}, of which some statistics are shown in \autoref{tab:stats}. Note that MWE tokens and types are underrepresented in the silver and bronze data compared to the gold data.
This is because the gold data contains more manual corrections on the token level than the silver and bronze data.
For the example of multi-word expressions see 
\autoref{fig:data_sample}. 
In the shared task, participants were allowed to use the silver and bronze data, this would especially make sense in the case of data-hungry neural models, though there is no guarantee that those representations resemble the gold standard.

\begin{table}[t]
\centering
\caption{Statistics for the PMB release 2.2.0 and the shared task evaluation set.}
\label{tab:stats}
\begin{tabular}{ l | r  r  r  r  r  r }
    \toprule
    \bf Data splits & \bf Docs & \bf Tokens & \bf Word types & \bf MWE tokens & \bf MWE types \\
    \midrule
    PMB 2.2.0 gold train & 4,597 & 29,195 & 4,431 & 789  & 494  \\
    PMB 2.2.0 gold dev & 682 & 4,067 & 1,251   & 71  & 61 \\
    PMB 2.2.0 gold test & 650 & 4,072 & 1,240   & 108  & 100  \\
    PMB 2.2.0 silver & 67,965 & 583,835  & 19,432   & 4,495  & 1,902  \\
    PMB 2.2.0 bronze & 120,662 & 919,247  & 23,354   & 2,054  & 699 \\ 
    \midrule
    Evaluation set & 600 & 4,066 & 1,237   & 92  & 79  \\
    \bottomrule
\end{tabular}
\vspace{-0.4cm}
\end{table}

The data provided to the shared task participants consists of pairs of a raw natural language text and its corresponding scoped meaning representation in clausal form.%
\footnote{\url{https://github.com/RikVN/DRS_parsing/tree/master/data/pmb-2.2.0}}
Whether the meaning representation is of gold, silver or bronze standard is explicitly indicated.
To facilitate automatic learning of scoped meaning representations, we also provided automatically induced alignments between clauses and tokens, where token positions are provided with character offsets. 
The examples of clause-token alignments are give in \autoref{fig:drs:play_piano} and \autoref{fig:data_sample}. 
The latter represents an exact formatting of the text and  clausal form pair provided in the shared task.

\begin{figure}[hbtp]
\centering
\begin{tabular}{c}
\begin{lstlisting}[basicstyle={\ttfamily\small}, escapechar=\$]
Nick Leeson was arrested for collapse of Barings Bank PLC.

b1 REF x1                      % Nick$\mytilde$Leeson [0...11]
b1 Name x1 "nick$\mytilde$leeson"       % Nick$\mytilde$Leeson [0...11]
b1 male "n.02" x1              % Nick$\mytilde$Leeson [0...11]
b2 REF t1                      % was [12...15]
b2 TPR t1 "now"                % was [12...15]
b2 Time e1 t1                  % was [12...15]
b2 time "n.08" t1              % was [12...15]
b2 REF e1                      % arrested [16...24]
b2 Patient e1 x1               % arrested [16...24]
b2 arrest "v.01" e1            % arrested [16...24]
b2 Theme e1 x2                 % for [25...28]
b2 REF x2                      % collapse [29...37]
b2 collapse "n.04" x2          % collapse [29...37]
b2 Patient x2 x3               % of [38...40]
b3 REF x3                      %
b3 Name x3 "barings$\mytilde$bank$\mytilde$plc"  % Barings$\mytilde$Bank$\mytilde$PLC [41...57]
b3 company "n.01" x3           % Barings$\mytilde$Bank$\mytilde$PLC [41...57]
                               % . [57...58]
\end{lstlisting}
\end{tabular}
\vspace{-0mm}
\caption{A sample of a training document (\href{http://pmb.let.rug.nl/explorer/explore.php?part=57&doc_id=0762}{57/0762}).
For each document there is a pair of raw text and the corresponding clausal form.
Clausal forms incorporate automatically induced clause-token alignment.}
\label{fig:data_sample}
\end{figure}


\subsection{Evaluation set}
\label{ssec:evaluation_set}

The official evaluation set contains 600 instances that were not released previously. They will not be released publicly, but are still available for (blind) scoring via the shared task website.%
\footnote{\url{https://competitions.codalab.org/competitions/20220}}
However, during the evaluation phase, we asked the participants to provide DRSs for a set of 12,606 short texts.
In addition to the raw texts (600) from the evaluation split, this set contained the train (4,597), development (682), and test (650) data from the PMB-2.2.0 release and the sentences (6,077) from the SICK dataset \citep{sick:14}.   
The reason for providing the inflated set of raw texts was three-fold: 
(i) Disguise the raw texts of the evaluation set to make it  hard to tune models on them;
(ii) Obtain the complete information about the performance of the systems on the provided training, development and test sets;
(iii) Carry out extrinsic evaluation of the participant systems on the natural language inference task.

\section{Evaluation Metrics and Baselines}
\label{sec:evaluation}

Before comparing a system produced clausal form to the gold one, the produced form is checked on validity---whether it represents a DRS. If the clausal form is invalid, it is replaced by a single non-matching clause.
In the shared task, we include three baseline systems.
The evaluation and validation scripts and the baselines are publicly available.%
\footnote{\url{http://github.com/RikVN/DRS_parsing}}

\subsection{Validation}
\label{ssec:referee}

Not all sets of clauses correspond to a well-formed DRS, e.g., discourse referents found in the conditions should be explicitly introduced in the boxes, or there should exist labelled boxes for the labels used in the discourse conditions.
We employ the validator \textsc{referee} \citep{drstacl:18} to automatically check a set of clauses on well-formedness.
\textsc{referee} does several checks for validity checking.
For example, first it scans each clause separately in a clausal form and identifies the types of variables based on the operators.
For each discourse referent variable, it checks the existence of the binding discourse referent.
During this procedure, \textsc{referee} also detects positions of the boxes in the DRS (i.e., so-called the subordinate relation).
Based on this information, it is checked that nested boxes do not create loops and there is a unique main box in the DRS.

All the released clausal forms of the DRSs are valid.
We provided the participants with \textsc{referee} in order to help them identify the ill-formed clausal forms produced by their systems.

\subsection{Evaluation}
\label{ssec:referee}

The evaluation defines to what degree a system output clausal form is similar to the corresponding gold one.
To compare the system output and gold representations, we compute the F1-score over the clauses, following \citet{allen:step2008}. 
We use the tool \textsc{counter} \citep*{lrec:2018}, which is specifically designed to evaluate DRSs. It is based on the \textsc{smatch} \cite{smatch:13} tool that is used to evaluate AMR parsers. It is essentially a hill-climbing algorithm that finds the best variable mapping between the produced DRS and the gold standard. To avoid local optima, we restart the procedure 10 times. In order to prevent an inflated F-score, before searching the maximal matching, \textsc{counter} discards those {\tt REF}-clauses which are deemed redundant.
A {\tt REF}-clause \mbox{$\langle$\ntt[0]{b REF x}$\rangle$} is redundant if and only if its discourse referent \ntt[0]{x} occurs with a concept predicate in a basic condition of the same box \ntt[0]{b} -- in other words, there exists a clause of the form \mbox{$\langle$\ntt[0]{b $concept$ "$pos.nn$" x}$\rangle$}.\footnote{In \autoref{fig:drs-match} redundant {\tt REF}-clauses are stricken through.}

\begin{figure*}[tph!]
\centering
\begin{tabular}{c c c}
Sample system output & Optimal mapping & Gold representation \\
\textsmaller[1]{\texttt{
\begin{tabular}[t]{@{\,}l@{\,}}\toprule
\matched{\drgvar{b3} IMP \drgvar{b2} \drgvar{b1}}\\
\strout{\drgvar{b2} REF \drgvar{x1}}\\
\nonmatched{\drgvar{b2} every "\posn{n}{01}" \drgvar{x1}}\\
\strout{\drgvar{b1} REF \drgvar{x2}}\\
\nonmatched{\drgvar{b1} Agent \drgvar{x2} \drgvar{x1}}\\
\matched{\drgvar{b1} new "\posn{a}{01}" \drgvar{x2}}\\
\matched{\drgvar{b1} Time \drgvar{x2} \drgvar{x3}}\\
\strout{\drgvar{b0} REF \drgvar{x3}}\\
\nonmatched{\drgvar{b0} time "\posn{n}{08}" \drgvar{x3}}\\
\nonmatched{\drgvar{b3} REF \drgvar{x0}}\\
\bottomrule
\end{tabular}
}}
&
\textsmaller[1]{\texttt{
\renewcommand\arraystretch{1.2}
\begin{tabular}[t]{l}
\noalv{2mm}
\drgvar{b3} $\mapsto$ \drgvar{b0}\\
\drgvar{b2} $\mapsto$ \drgvar{b1}\\ 
\drgvar{b1} $\mapsto$ \drgvar{b2}\\
\drgvar{b0} $\mapsto$ \drgvar{b3}\\  
\drgvar{x1} $\mapsto$ \drgvar{x1}\\
\drgvar{x2} $\mapsto$ \drgvar{s1}\\
\drgvar{x3} $\mapsto$ \drgvar{t1}\\
\drgvar{x0} $\mapsto$ \textnormal{\textsmaller[1]{N/A}}
\end{tabular}
}}
&
\textsmaller[1]{\texttt{
\begin{tabular}[t]{@{\,}l@{\,}}\toprule
\matched{\drgvar{b0} IMP \drgvar{b1} \drgvar{b2}}\\
\strout{\drgvar{b1} REF \drgvar{x1}}\\
\nonmatched{\drgvar{b1} entity "\posn{n}{01}" \drgvar{x1}}\\
\strout{\drgvar{b2} REF \drgvar{s1}}\\
\nonmatched{\drgvar{b2} Theme \drgvar{s1} \drgvar{x1}}\\
\matched{\drgvar{b2} new "\posn{a}{01}" \drgvar{s1}}\\
\matched{\drgvar{b2} Time \drgvar{s1} \drgvar{t1}}\\
\strout{\drgvar{b3} REF \drgvar{t1}}\\
\nonmatched{\drgvar{b0} time "\posn{n}{08}" \drgvar{t1}}\\
\nonmatched{\drgvar{b0} EQU \drgvar{t1} "now"}\\
\bottomrule
\end{tabular}
}}
\end{tabular}
\\[3mm]

\hspace{-2mm}
\begin{tabular}{@{}cc@{}}
\color{gray!80}
\begin{drstree}{-1}{3mm}{3mm}
[{\pdrs{b3}{$x_0$}{\noalv{-6pt}
    \pdrs{b2}{$x_1$}{
        every$\sym{.n.01}(x_1)$} 
    \raisebox{-13pt}{$\Rightarrow$}
    \pdrs{b1}{$x_2$}{
        new$\sym{.a.01}(x_2)$\\
        \wsp$\sym{Time}(x_2, x_3)$\\
        \wsp$\sym{Agent}(x_2, x_1)$}}}
    [{\pdrs{b0}{$x_3$}{
        $\sym{time.n.08}(x_3)$}}
    ]
]            
\end{drstree}
&
\color{gray!80}
\begin{drstree}{-1}{3mm}{3mm}
[{\pdrs{b0}{$t_1$}{$\sym{time.n.08}(t_1)$\\
   \wsp$t_1 = \sym{now}$\\
   \noalv{-6pt}
    \pdrs{b1}{$x_1$}{
        entity$\sym{.n.01}(x_1)$} 
    \raisebox{-13pt}{$\Rightarrow$}
    \pdrs{b2}{$s_1$}{
        new$\sym{.a.01}(s_1)$\\
        \wsp$\sym{Time}(s_1, t_1)$\\
        \wsp$\sym{Theme}(s_1, x_1)$}}}
]            
\end{drstree}
\end{tabular}
\caption{An optimal mapping of variables which maximizes overlap between the system output and gold clausal forms for the sentence (PMB document \href{http://pmb.let.rug.nl/explorer/explore.php?part=00&doc_id=2302&type=der.xml}{00/2302}) \emph{Everything is new}.
The maximal overlap yields an F-score of $42.9$.
Matching, non-matching and redundant clauses are in green, red, and stricken through, respectively. 
The box-notation of scoped meaning representations is not available during the comparison of clausal forms.} 
\label{fig:drs-match}
\vspace{-0.4cm}
\end{figure*}

An example of comparing the clausal forms of two scoped meaning representations is shown in \autoref{fig:drs-match}. 
With respect to the optimal mapping, both, the sample system output and gold clauses, include three clauses that could not be matched with each other while four clauses are matched. 
The optimal mapping gives us a precision and recall of $\nicefrac{3}{7}$, resulting in an F-score of $42.9$. Similarly to AMR, we use micro-averaged F-score when evaluating a set of DRSs.

An aspect that is different from the AMR evaluation system is that we generalize over synonyms. 
In a preprocessing step of the evaluation, all word senses are converted to its WordNet 3.0 synset ID. 
For example, fox\sym{.n.02} and dodger\sym{.n.01} both get normalized to dodger\sym{.n.01} and are thus able to match.

To calculate whether two systems differ significantly, we perform approximate randomization \cite{random-appr:89}, with $\alpha$ = $0.05$, \emph{R} = $1000$ and $F(model_{1}) > F(model_{2})$ as test statistic for each individual DRS pair. 

\subsection{Baselines}

We provide three baseline parsers: \textsc{spar}, \textsc{sim-spar} and \textsc{amr2drs}. \textsc{spar} simply outputs a default DRS, which is a DRS that is the most similar to the DRSs in our training set.\footnote{For PMB release 2.2.0 this is the DRS for \emph{Tom voted for himself.}} 
\textsc{sim-spar} outputs the DRS of the most similar sentence in the training set, based on the cosine distance of the average word-embedding vector, calculated using GloVe \citep{glove:14}.
\textsc{amr2drs} is a script that converts the output of an AMR parser to a valid DRS by applying a set of rules, described in \citet{bos:16} and \citet*{lrec:2018}. 
We will provide scores on the development, test and evaluation sets by using the AMR parser of \citet{clinAMR:17}.

\section{Participating Systems}

We received a total of five submissions in the shared task out of 32 registered participants.
Three out of five submitted a system paper.
The general characteristics of the participating systems are give in \autoref{tab:systems}. 
Following \citet{nissim2017sharing}, we explicitly encouraged the participants to include ablation experiments and negative results (if any).
Note that the authors of the systems \taclsys{} and \iwcssys{} are from the organizers.
Below, we provide a short description of each system.

\begin{table}[hbtp]
\centering
\caption{Overview of the participating systems}
\label{tab:systems}
\begin{tabular}{llcccc}
\toprule
                            & \textbf{Model} & \textbf{Input} & \textbf{Embeddings} & \textbf{Silver} & \textbf{Bronze} \\
\midrule
\liusys      & Transformer    & char       &   \cross         &   \cmark             & \cmark     \\
\iwcssys     & seq2seq        & char        &   \cross      &     \cmark                 &  \cross                    \\
\taclsys     & seq2seq        & char        &   \cross       &  \cmark                    & \cross                     \\
\evang       & stack-LSTMs    & word        &   \cmark       &    \cross                  &   \cross                   \\
\fancellusys & bi-LSTM       & word         &    \cmark      &   \cross                   & \cross       \\
\bottomrule
\end{tabular}
\vspace{-0.4cm}
\end{table}

\subsection{\citet{drstacl:18}}

\taclsys{}, the parser described in \citet{drstacl:18}, uses a character-level neural sequence-to-sequence model to produce DRSs. They apply a number of methods to improve performance, such as rewriting the variables to a more general format, introducing a feature for uppercase letters and \emph{not} using character-level representation for DRS roles and operators. Moreover, they show that performance can be substantially improved by first pre-training on gold and silver data, after which the parser is fine-tuned on only the gold standard data.
\vspace{-0.1cm}
\subsection{\citet{W19:noord}}
The system of \iwcssys{} is the parser described in \citet{drsiwcs:19} and \citet{W19:noord}, which follows up on their work previously described in \citet{drstacl:18}. They improve on this work in two ways: (i) by switching their sequence-to-sequence framework from OpenNMT \citep{opennmt:17} to Marian \citep{mariannmt} and (ii) by providing the encoder with linguistic information (lemmas, semantic tags \citep{semantic-tagset:17}, POS-tags, dependency parses and CCG supertags) that are encoded in a separate encoder.
\vspace{-0.1cm}

\subsection{\citet{W19:liu}}
\liusys{} also follow the approach of \citet{drstacl:18} in terms of pre- and postprocessing the data, but they improve on it by using the Transformer model \citep{vaswani2017attention}, instead of a sequence-to-sequence RNN. Also, they show that employing the bronze standard in addition to the gold and silver standard leads to improved performance.
\vspace{-0.1cm}

\subsection{\citet{W19:evang}}
\evangsys{} aim to find a middle-ground between traditional symbolic approaches and the recent neural (sequence-to-sequence) models. They employ a transition-based parser that relies on explicit word-meaning pairs that are found in the training set. Parsing decisions are made based on vector representations of parser states, which are encoded using stack-LSTMs.
\vspace{-0.1cm}

\subsection{\fancellusys\protect\footnotemark}
\footnotetext{The full list of authors: Federico Fancellu, Sorcha Gilroy, Adam Lopez and Mirella Lapata (University of Edinburgh).
Since the authors refrained from submitting a system paper due to the ACL policy for submission, we include a slightly extended summary of their system, generously provided by them.}
\fancellusys{} propose a graph decoder that given an input sentence encoded via a bidirectional LSTM generates a DAG (Directed Acyclic Graph) as a sequence of fragments from a graph grammar. 
These fragments are \textit{delexicalized}; predicate names, synset and information on whether the predicate is presupposed or not are predicted in a second step, conditioned on the fragment and the decoding history.  
Two are the main features of the graph parser: 1) it is agnostic to the underlying semantic formalism and does not need any preprocessing step to deal with variable binding; 
2) fragments are aware of the overall graph structure and the graph is built incrementally via a process of non-terminal rewriting. 
(1) sets this method apart from the graph parser of \citet{groschwitz2018amr} where a grammar is extracting via an elaborate pre-processing step, tailored to a specific formalism, whereas (2) allows to leverage neural sequential decoding \citep[stackLSTM,][]{dyer2016recurrent}.
The only preprocessing step required is to convert DRSs in clause format into single-rooted, fully instantiated DAGs; 
we do so by treating both variables and boxes as nodes and semantic roles, operators and discourse relations as edges between those (where each binary operator or relation gives rise to two edges). 
Similarly, the only postprocessing step lies in converting the graph back to clause format. This last step can inject errors in the parse and it is the reason why some of the output graphs can be ill-formed.

\section{Results}

\begin{table}[t]
\centering
\caption{Official results of the shared task for the participating systems}
\begin{tabular}{ l | c  c  c | c  c  c }
    \toprule
        & \multicolumn{3}{c|}{PMB 2.2.0 (F\%)} & \multicolumn{3}{c}{Evaluation set (\%)}\\
     & Train & Dev & Test & Prec. & Rec. & F \\ 
     \midrule
     \textsc{amr2drs}  & NA & 39.7 & 40.1 & \unimp{36.7} & \unimp{42.2} & 38.8 \\
     \textsc{spar} & NA & 40.0 & 40.8 & \unimp{44.3} & \unimp{35.4} & 39.4 \\
     \textsc{sim-spar}  & NA & 53.3 & 57.7 & \unimp{55.7} & \unimp{53.0} & 54.3 \\
     \midrule
    \fancellusys& 91.1 & 69.9 & 73.3 & \unimp{71.9} & \unimp{64.1} & 67.8 \\
    \evangsys  & 84.2 & 74.4 & 74.4 & \unimp{71.9} & \unimp{69.9} & 70.9 \\
    \taclsys{}  & 88.5 & 81.2 & 83.3 & \unimp{80.8} & \unimp{78.6} & 79.7 \\
    \iwcssys{} & 94.9 & 86.5 & 86.8 & \unimp{85.5} & \unimp{83.6} & 84.5 \\
    \liusys     & 96.9 & 85.5 & 87.1 & \unimp{84.8} & \unimp{84.8} & 84.8 \\
    \bottomrule
\end{tabular}
\label{tab:results}
\end{table}

\autoref{tab:results} shows the official results of the shared task. The system of \liusys \ achieved the best performance, though there is no significant difference with the work of \iwcssys \ ($p = 0.23$). The systems of \evangsys \ and \fancellusys , though clearly outperforming the baselines, are a bit behind the best three systems. However, they can likely improve performance by incorporating silver and bronze standard data. No systems seem to have overfit on the provided dev and test sets. The work of \fancellusys \ is perhaps overfit on the training set, given their high score on train compared to the test sets.

\autoref{tab:auto_eval} shows a more detailed overview of the results. All teams produced a substantial amount of perfect DRSs, but only 31 DRSs of them were perfectly produced by each system.
\evangsys{} is the only system with a substantial number of ill-formed DRSs. This hurts their performance, since they get an F-score of 0.0 in evaluation. If we ignore referee and score their ill-formed DRSs as if they were valid, their score increases to 72.1. On the other hand, calculating an F-score for \emph{only} the ill-formed DRSs (without referee) gives us an F-score of $50.6$, suggesting that the model would not have scored very well in either way.

Similar as was observed in \citet{drstacl:18}, word sense disambiguation is problematic for the DRS parsers. When assuming oracle sense numbers, all systems obtain a substantially higher F-score (increases of $1.8$ to  $3.6$). \iwcssys{} propose a simple method to improve on this sub-problem by taking the most frequent sense in the training set for a concept, though this only increased their F-scores by $0.2$ to $0.4$ on the dev and test sets. Nouns are the easiest for all models to correctly produce (possibly due to the frequent \texttt{time.n.08}), while adverbs are the hardest, though there are only 12 such clauses in the evaluation set.

\begin{table}[t]
\centering
\caption{F-scores of fine-grained evaluation of the participating systems on the evaluation set. ``Winner out of 5'' counts only those instances for which the parser obtained a higher score than all the rest, while ``Highest out of 5'' allows ties.}
\begin{tabular}{lccccc}
\toprule
                          & \liu{} & \iwcs{} & \tacl{} & \evang{} & \fancellu{}  \\ \midrule
\textbf{All clauses}  & 84.8 & 84.5 & 79.7 & 70.9 & 67.8          \\ \midrule
\textbf{DRS Operators}  &   93.9 & 94.2 & 91.7 & 75.2 & 76.3           \\
\textbf{VerbNet roles}   &     82.7 & 83.5 & 78.1 & 72.4 & 66.4    \\
\textbf{WordNet synsets} & 83.8 & 82.3 & 77.2 & 67.8 & 66.5            \\
\textbf{\quad nouns}  & 89.2 & 87.5 & 83.5 & 75.9 & 70.3      \\ 
\textbf{\quad verbs}  & 69.5 & 68.9 & 60.9 & 44.1 & 58.3         \\ 
\textbf{\quad adjectives}  & 74.8 & 74.2 & 66.5 & 61.5 & 53.8            \\ 
\textbf{\quad adverbs} & 63.6 & 45.5 & 33.3 & 0.0 & 31.6       \\ \midrule
\textbf{Oracle sense numbers} & 86.6 & 87.1 & 82.6 & 74.5 & 69.8           \\
\textbf{Oracle synsets} & 90.5 & 90.7 & 87.5 & 80.1 & 76.5            \\
\textbf{Oracle roles} & 88.4 & 88.5 & 84.3 & 74.5 & 73.7   \\  
 \midrule
 \textbf{\# of perfect DRSs} & 214 & 210  & 160   & 95  & 104 \\
 \textbf{\# highest out of 5} & 383 & 376 & 261 & 171 & 161  \\
 \textbf{\# winner out of 5} & 100  & 77  & 26  & 18  & 18   \\
 \textbf{\# of ill-formed DRSs} & 1 & 0 & 1 & 37 & 5  \\
 \bottomrule 
\end{tabular}
\label{tab:auto_eval}
\end{table}

\section{Analysis}

\begin{figure}[!t]
  \centering
  \includegraphics[width=0.75\textwidth]{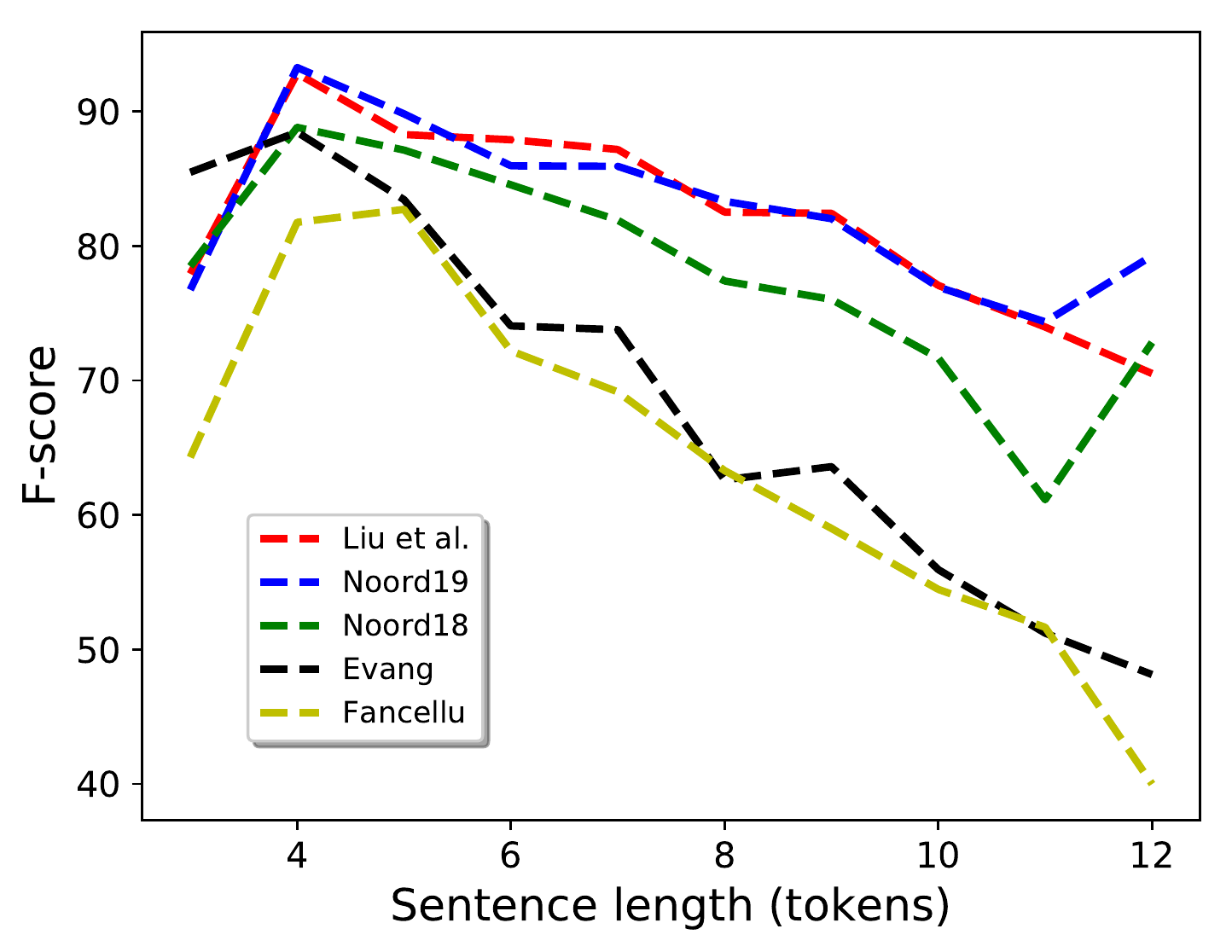}
  \caption{\label{fig:senlen}Performance of the systems per sentence length (punctuation are counted as tokens).}
\end{figure}

Longer sentences are probably harder to parse, but which systems behave well on longer sentences?
\autoref{fig:senlen} shows the performance of the systems plotted over sentence length. As expected, all systems show a clear drop in performance for longer sentences. The work of \liusys{} is based on the Transformer model \citep{vaswani2017attention}, which claims that performance should not degrade for longer sentences. However, for DRS parsing this does not seem to be the case, as \liusys{} shows a similar decrease in performance as the neural models of \taclsys{} and \iwcssys{}.

How similar were the outputs of the participating systems to each other? 
\autoref{tab:comp} shows pairwise comparison of the outputs of the systems on the evaluation set. 
The only system that has a substantially higher similarity to one of the systems than their official F-score is \taclsys{} compared to \iwcssys{}, which tells us the models make similar mistakes. This makes sense given that they are both character-level sequence-to-sequence models trained on the same data.
Additionally, the output of \iwcssys{} comes closest to the output of \fancellusys{} when compared to other systems' outputs.
Similarly, \evangsys{} is most similar to \taclsys{} than to any other systems.

\begin{table}[t]
\centering
\caption{\label{tab:comp}F-scores of all systems compared to each other.}
\begin{tabular}{l|ccccc}
\toprule
          & ~ ~~\liu~~ ~ & \iwcs & \tacl & ~~\evang~~ & \fancellu\\\midrule
\liu      & \clcl{} &   83.4&   79.6 & 70.5  &      66.6          \\
\iwcs     &   83.4  &\clcl{}&   83.3       & 70.9    & 68.1      \\
\tacl     &   79.6  &  83.3 & \clcl{} & 71.5        & 67.1       \\
\evang    &   70.5  &  70.9 &  71.5   & \clcl{}  &  61.3       \\
\fancellu &   66.6  &  68.1        &   67.1       &  61.3 & \clcl{} \\
\bottomrule
\end{tabular}
\end{table}

How complementary were the participating systems to each other? If we had an ensemble system, with an oracle component that selected the best DRS for each sentence out of the participants submissions, it would obtain an F-score of 90.5. When only combining the submissions of \liusys{} and \iwcssys, it would already result in an F-score of 89.1. This suggests that the neural models in fact do learn different things, though there is still a significant portion that both methods could not learn.

Finally, are there phenomena that are especially hard for all participating systems? This is not an easy question to answer, and here we show just a first step to such an analysis.
\autoref{tab:worst} shows sentences for which systems, on average, performed badly. Some of them show non-standard use of English, others are phenomena that are relatively rare, such as generics, multi-word expressions, and coordination.

\begin{table}[htb]
\centering
\caption{Sentences for which participating systems, on average, produced the worst DRSs}
\label{tab:worst}
\begin{tabular}{l@{}cl}
\toprule
\textbf{Sentence} & \textbf{avg. F} & \textbf{Comment}\\
\midrule
Thou speakest. & 21.4 & archaic English\\
I dinnae ken. & 21.8 & Scottish\\
My fault. & 24.2  & noun phrase\\
A cat has two ears. & 38.1 & generic\\
I look down on liars and cheats. & 40.3 & coordination, MWE\\
Get me the number of this young girl. & 41.8 & imperative\\
She attends school at night. & 44.6 & temporal modifier \\
The union of Scotland and England took place in 1706. & 46.4 & coordination, MWE\\
Something I hadn't anticipated happened. & 47.0 & reduced relative clause\\
Charles I had his head cut off. & 47.2 & ordinal, MWE\\
\bottomrule
\end{tabular}
\end{table}

\section{Conclusion}

The first shared task on DRS parsing was successful. It improved the state-of-the-art in DRS parsing, and the variety in methods used (models based on recursive neural networks, transformer models, models based on transition-based parsing, graph decoders) gives inspiration for future research.  In the future DRS parsing will be made more challenging by moving to longer sentences and texts (where we expect simple seq2seq models to have a harder time), more complex phenomena (ellipsis, comparatives, multi-word expressions), and to languages other than English. 

\section*{Acknowledgements}

We would like to thank the IWCS-2019 organizers, Stergios Chatzikyriakidis, Vera Demberg, Simon Dobnik
and Asad Sayeed, for hosting the shared task in Gothenburg.
We also would like to thank all participants for their feedback on the task. 
This work was funded by the NWO-VICI grant ``Lost in Translation -- Found in Meaning'' (288-89-003).

\bibliography{mybib}
\bibliographystyle{chicago}

\end{document}